\documentclass[11pt]{article}

\usepackage[preprint]{acl}

\usepackage{times}
\usepackage{latexsym}

\usepackage[T1]{fontenc}

\usepackage[utf8]{inputenc}

\usepackage{microtype}

\usepackage[utf8]{inputenc} 
\usepackage{hyperref}[]
\usepackage{url}            
\usepackage{booktabs}
\usepackage{natbib}
\usepackage{amsfonts}      
\usepackage{nicefrac}       
\usepackage{microtype}      
\usepackage{amsmath}
\usepackage{amssymb}
\usepackage[section]{placeins}
\usepackage{graphicx}
\usepackage{caption}
\usepackage{natbib}
\usepackage{booktabs}
\usepackage{colortbl} 
\usepackage{soul} 
\usepackage{tabularx}
\usepackage{hyperref}
\usepackage{stfloats}
\usepackage{algorithm,algorithmicx}
\usepackage[noend]{algpseudocode}
\usepackage{algcompatible}
\graphicspath{{./figures/}}
\usepackage{multirow}
\usepackage{mathtools}
\usepackage{shortvrb}
\usepackage{xspace}
\usepackage{dsfont}
\usepackage[most]{tcolorbox}
\usepackage[normalem]{ulem}
\useunder{\uline}{\ul}{}

\usepackage{amsmath,amsfonts,bm}

\def\eqref#1{equation~\ref{#1}}

\def\1{\bm{1}}

\DeclareMathAlphabet{\mathsfit}{\encodingdefault}{\sfdefault}{m}{sl}
\SetMathAlphabet{\mathsfit}{bold}{\encodingdefault}{\sfdefault}{bx}{n}

\definecolor{lightgrey}{HTML}{dcdbdb}
\definecolor{lightblue}{HTML}{E8F0FE}
\definecolor{lin}{HTML}{A810FE}
\definecolor{gray}{HTML}{9aa0a6}
\definecolor{lightpink}{HTML}{F48FB1}
\definecolor{lightred}{HTML}{FFCBC9}
\definecolor{lightcyan}{HTML}{80DEEA}

\newtcolorbox{mybox}[2][]
  {colback = black!5!white, colframe = black!75!black, fonttitle = \bfseries,
    colbacktitle = black!100!black, enhanced, 
    attach boxed title to top left={yshift=-2.2mm,xshift=4mm},
    title=#2,#1}

\usepackage{inconsolata}

\usepackage{graphicx}

\usepackage{svg} 

\title{DreamReasoner-8B: Block-Size Curriculum Learning for \\Diffusion Reasoning Models}

\newcommand{\ourmodel}{DreamReasoner\xspace}

\author{
    Zirui Wu$^{1,2}$,
    Lin Zheng$^{1}$,
    Jiacheng Ye$^{1}$,
    Shansan Gong$^{1}$,
    Xueliang Zhao$^{1}$,\\
    \textbf{Yansong Feng}$^{2}$,
    \textbf{Wei Bi},
    \textbf{Lingpeng Kong}$^{1}$\thanks{\;\;Corresponding author.}\\
    $^1$The University of Hong Kong
    $^2$Peking University\\
    {\tt ziruiwu@pku.edu.cn}, 
    {\tt lpk@cs.hku.hk}
}

\begin{document}
\maketitle

\begin{abstract}

Block diffusion language models accelerate decoding through parallel block-wise denoising, yet whether they can be reliably scaled for long chain-of-thought (CoT) reasoning remains unresolved. To this end, we develop \ourmodel-8B, an open-source block diffusion reasoning model, and conduct a systematic study of how training and inference block sizes affect long-CoT reasoning. Our analysis reveals a stark performance disparity: training with large block sizes yields remarkably poor reasoning, whereas small block sizes preserve effective reasoning. To bridge this granularity gap, we propose block-size curriculum learning, which gradually transitions training from fine-grained to coarse-grained block sizes, thereby overcoming this limitation and enabling strong reasoning performance that generalizes across diverse inference block sizes. On mathematical and code reasoning benchmarks, \ourmodel-8B achieves results competitive with leading open autoregressive models such as Qwen3-8B. This work establishes a practical foundation for efficient, reasoning-capable diffusion language models. We release our model at \url{https://github.com/DreamLM/DreamReasoner}.\looseness=-1

\end{abstract}
\section{Introduction}
\label{sec:intro}

Autoregressive (AR) language models have become the dominant paradigm for complex reasoning, owing to their strong sequential coherence and the demonstrated ability to generate extended chains of thought \citep{openai2026gpt54, deepseekai2026deepseekv4,kimiteam2026kimik25visualagentic}. Yet the strict left-to-right factorization in AR decoding fundamentally limits parallelism during inference. This efficiency bottleneck has motivated increasing interest in alternative architectures that can preserve reasoning quality while enabling greater generation parallelism.\looseness=-1

\begin{figure*}[t]
    \centering
    \begin{minipage}[c]{0.26\linewidth}
        \centering
        \includegraphics[width=\linewidth]{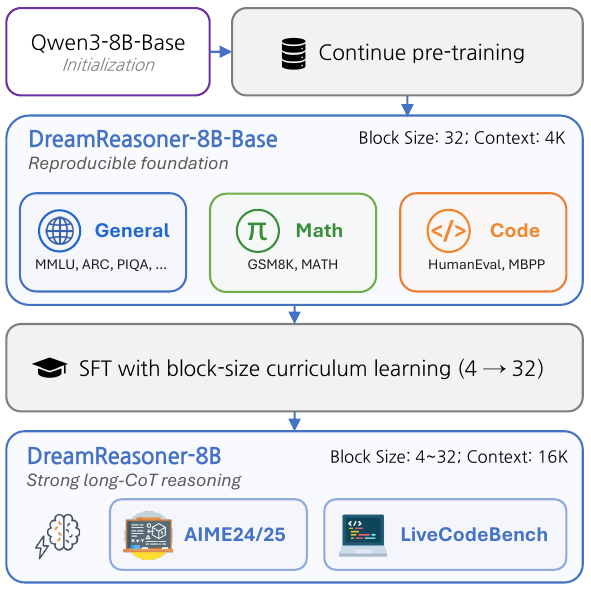}
    \end{minipage}
    \hspace{0.02\linewidth}
    \begin{minipage}[c]{0.69\linewidth}
        \centering
        \includegraphics[width=\linewidth]{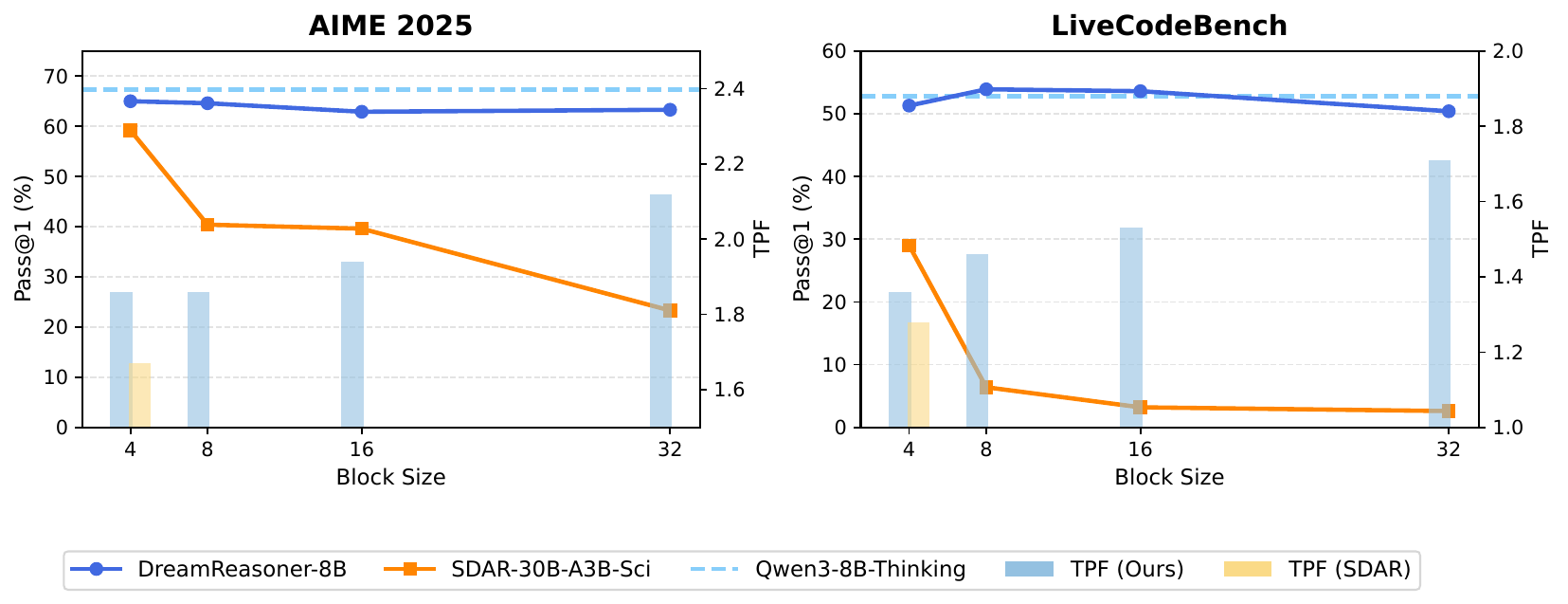}
    \end{minipage}
    \caption{
    Overview and performance of \ourmodel-8B.
    \textbf{Left:} the training pipeline of \ourmodel-8B-Base and \ourmodel-8B through block-size curriculum learning.
    \textbf{Right:} \ourmodel-8B maintains high accuracy close to the Qwen3-8B-Thinking baseline while its efficiency (measured by TPF\protect\footnotemark{}) improves with larger block sizes. In contrast, SDAR-30B-A3B-Sci shows lower efficiency and its accuracy degrades sharply as the block size increases, highlighting the superior balance between performance and computational cost achieved by our model.\looseness=-1}
    \label{fig:overview}
\end{figure*}

Diffusion language models have emerged as a promising alternative, offering parallel token generation, bidirectional context modeling, and flexible decoding strategies \citep{sahoo2024simple, gong2025diffusionllm, nie2025llada, ye2025dream}. Among these, block diffusion \citep{arriola2025blockdiffusion, bie2025llada20scalingdiffusionlanguage,bie2026llada21speedingtextdiffusion} provides a principled structural solution that interpolates between pure diffusion and pure autoregressive generation. By partitioning the output sequence into contiguous blocks, block diffusion handles inter-block dependencies autoregressively and intra-block tokens in parallel via bidirectional diffusion denoising, thereby enabling a flexible trade-off between sequential coherence and generation parallelism. Nevertheless, open-source reasoning-capable diffusion models still lag behind AR counterparts on standard benchmarks \citep{wang2025revolutionizing, cheng2025sdar}, and these models typically exhibit substantial performance degradation as block size increases as shown in Figure~\ref{fig:overview}, undermining the efficiency gains promised by coarse-grained block decoding.

In this work, we study whether larger blocks can be made useful for reasoning in diffusion language models. Specifically, we investigate whether block diffusion models can be trained to reason effectively with large block sizes, and identify the training and decoding strategies that enable this. To answer this question, we train \ourmodel-8B, an open 8B-parameter block-diffusion reasoning model, and systematically evaluate how training block size, inference block size, and decoding strategy interact. Rather than treating block size as a fixed architectural constant, we analyze it as a \textit{scaling axis} that jointly affects reasoning quality and decoding efficiency.

Our analysis reveals several key findings:
First, naively training with a large block size leads to degradation in reasoning performance while small block sizes preserve reasoning capabilities.
Second, block-size curriculum learning that gradually transitions training from fine-grained to coarse-grained block sizes can overcome the degradation associated with large-block training.
Third, curriculum-trained models achieve substantial efficiency gains from large-block inference without sacrificing reasoning quality. On mathematical and code reasoning benchmarks, \ourmodel-8B achieves performance competitive with leading open autoregressive models such as Qwen3-8B-Thinking~\citep{yang2025qwen3}, substantially outperforming prior block diffusion models of comparable scale.

Nevertheless, we find that the gain in decoding efficiency remains fundamentally bottlenecked by conservative token-level confidence thresholds during decoding. To analyze this bottleneck, we introduce RelaxedConfidence, an analytical decoding probe that examines how contiguous spans of high-confidence tokens can be committed earlier through relaxed thresholds. This yields average TPF (tokens per forward pass) gains of 22.5\% and 54.5\% in the thinking and answering phases without compromising reasoning fidelity.\looseness=-1

\footnotetext{We regard the TPF of autoregressive models as 1. For SDAR, we report TPF only at block size 4, since its accuracy degrades substantially at larger block sizes.}

Our contributions are summarized as follows:
\begin{itemize}
    \item We systematically investigate block size as a scaling axis for diffusion reasoning models, demonstrating that large-block training is brittle at scale, whereas curriculum learning substantially stabilizes large-block generation.
    
    \item We release \ourmodel-8B, an open block-diffusion reasoning model that achieves competitive reasoning performance on par with Qwen3-8B-Thinking.
    
    \item We introduce RelaxedConfidence, an analytical decoding probe revealing that relaxing the commitment threshold for tokens surrounded by reliable neighbors enables earlier and broader parallel decoding.
\end{itemize}

\section{Preliminaries}
\label{sec:preliminary}
Full-sequence discrete diffusion processes jointly model all token positions in an input sequence. This provides bidirectional context and flexible decoding orders, but it also couples denoising across the entire sequence, which can slow training convergence on large corpora~\citep{ni2025diffusion,prabhudesai2025diffusion} and make long-form generation vulnerable to error accumulation. Block diffusion mitigates these issues by combining autoregressive sequence decomposition with local parallel denoising, enabling scalable training and efficient inference~\citep{han-etal-2023-ssd,arriola2025blockdiffusion,bie2025llada20scalingdiffusionlanguage,cheng2025sdar,fu2025efficientdlm}.

Let $\mathbf{x}_0$ denote the clean input sequence, and partition it into $K$ contiguous blocks $\mathbf{x}_0 = (b^1_0, b^2_0, \dots, b^K_0)$, each containing $B$ tokens. We use superscripts to index blocks and subscripts to index diffusion time steps; thus $b^k_t=(x^{k,1}_t,\dots,x^{k,B}_t)$ denotes the state of the $k$-th block at time $t$, and $b^k_0$ is its clean state. Block diffusion preserves an autoregressive factorization across blocks while modeling each block conditional with a local bidirectional diffusion process:
\begin{equation}
    p_{\theta}(\mathbf{x}_0) = \prod_{k=1}^{K} p_{\theta}(b^k_0 \mid b^{<k}_0),
    \label{eq:block-joint}
\end{equation}
where $b^{<k}_0 \coloneqq (b^1_0,\dots,b^{k-1}_0)$. Each conditional $p_{\theta}(b^k_0 \mid b^{<k}_0)$ is parameterized by a local \emph{diffusion process} that processes all tokens within $b^k$ in parallel. The process uses an absorbing corruption procedure with a dedicated absorbing token \texttt{[MASK]}. For the $k$-th block, the forward process corrupts only the target block while keeping the prefix context clean. Generation reverses this process: starting from a fully masked block state, the model iteratively predicts $p_\theta(x^{k,i}_0\mid b^{<k}_0,b^k_t)$ for masked positions and refines them in parallel, conditioned on fixed preceding blocks $b^{<k}_0$ as context.

\paragraph{Training objective.}
The objective can be reduced to a weighted cross-entropy on the masked positions for each block~\citep{arriola2025blockdiffusion,cheng2025sdar,bie2025llada20scalingdiffusionlanguage}, conditioned on the clean previous context $b^{<k}_0$. Let $\alpha_t \in (0,1)$ denote the noise schedule, $\alpha_t' \coloneqq \frac{d\alpha_t}{dt}$, and $w_t \coloneqq \frac{\alpha_t'}{1-\alpha_t}$ denote the loss weight at time step $t$. Aggregating over all $K$ blocks yields:
\begin{equation}
\begin{split}
\mathcal{L}(\theta) \!=\!
-\mathbb{E}_{t,b_0,b_t}\biggl[\sum_{k=1}^{K}w_t
\log p_{\theta}\bigl(b^{k}_0 \!\mid\! b^{<k}_0\!,b^k_t\bigr)\biggr]\!,
\end{split}
\label{eq:block-loss}
\end{equation}
where $t \sim \mathcal{U}(0,1)$, $b_0=\{b^k_0\}_{k=1}^{K}$ represents the block sequence of the clean input data, and $b_t=\{b^k_t\}_{k=1}^{K}$ is its corrupted version at time $t$. For each block, only masked tokens contribute to the loss with their context provided by clean prefix blocks.\looseness=-1

\begin{table*}[tbp]
\centering
\caption{Base model performance comparison on standard benchmarks. Best results are shown in bold, and second-best results are underlined. }
\label{tab:dream-main}
\scalebox{0.75}{
\begin{tabular}{lccccccc}
\toprule
\textbf{Model} & \textbf{\ourmodel-8B} & \textbf{Dream-v0-7B} & \textbf{LLaDA-8B} & \textbf{Qwen3-8B} & \textbf{MiMo-7B} & \textbf{OLMo3-7B} & \textbf{Qwen2.5-7B} \\
\midrule
Type & Block Diffusion & Diffusion & Diffusion & AR & AR & AR & AR \\
\midrule
\multicolumn{8}{c}{\textit{General Tasks}} \\
\midrule
MMLU & \underline{75.4} & 69.5 & 65.9 & \textbf{76.9} & 69.8 & 64.5 & 71.9 \\
ARC-E & \textbf{87.0} & \underline{83.9} & 71.8 & 81.6 & 76.5 & 80.6 & 77.4 \\
ARC-C & \textbf{63.5} & \underline{59.8} & 47.5 & 56.9 & 51.6 & 53.5 & 51.5 \\
HellaSwag & 76.6 & 73.3 & 72.7 & \underline{78.6} & 77.5 & 74.1 & \textbf{78.9} \\
PIQA & \textbf{81.2} & 75.8 & 75.8 & 79.4 & \underline{79.9} & 78.6 & 79.8 \\
WinoGrande & 74.2 & 74.5 & 74.5 & \textbf{77.5} & 75.9 & 73.5 & \underline{76.1} \\
RACE & \underline{43.8} & \textbf{44.7} & 38.7 & 42.3 & 38.6 & 40.1 & 41.9 \\
\midrule
\multicolumn{8}{c}{\textit{Math \& Science}} \\
\midrule
GSM8K & \underline{83.4} & 77.2 & 70.9 & \textbf{86.5} & 73.8 & 76.5 & 78.9 \\
MATH & \textbf{55.8} & 39.6 & 30.7 & \underline{52.7} & 35.8 & 41.5 & 41.1 \\
GPQA & \textbf{42.9} & 36.6 & 30.4 & \underline{41.1} & -- & 36.6 & 35.5 \\
\midrule
\multicolumn{8}{c}{\textit{Code}} \\
\midrule
HumanEval & \textbf{69.5} & 57.9 & 32.9 & \underline{68.9} & 51.8 & 43.9 & 56.7 \\
MBPP & \underline{71.7} & 56.2 & 39.0 & \textbf{72.4} & 69.2 & 66.7 & 63.6 \\
\bottomrule
\end{tabular}}
\end{table*}

\section{\ourmodel-8B-Base}
\label{sec:pretraining}

Following previous practice~\citep{gong2025diffusionllm,ye2025dream,cheng2025sdar,fu2025efficientdlm}, we first initialize from Qwen3-8B-Base \citep{yang2025qwen3} and continually pre-train DreamReasoner-8B-Base as a block-diffusion model, establishing a reproducible foundation for subsequent fine-tuning. We use the block-diffusion loss objective in Eq.~\ref{eq:block-loss} with block size $32$ throughout continual pretraining.\looseness=-1

\paragraph{Data composition.}
We assemble a pretraining corpus by curating high-quality, open-source datasets used by open-source language models, including OLMo~3~\citep{olmo2025olmo3} and Nemotron Nano V3~\citep{nvidia2025nemotron3nano}. The data mixture is designed to enhance reasoning capabilities, and the final training corpus contains 160B tokens.

\paragraph{Baselines.} 
We compare \ourmodel-8B-Base against two categories of open-source baselines with settings listed in Appendix~\ref{app:impl_details}: (i) \textit{Autoregressive base models}, including Qwen3-8B-Base~\citep{yang2025qwen3}, MiMo-7B-Base~\citep{xiaomi2025mimo}, 
OLMo3-7B-Base~\citep{olmo2025olmo3}, and Qwen2.5-7B-Base~\citep{qwen2025qwen25technicalreport}; 
and (ii) \textit{Diffusion base models}, including the full-sequence 
diffusion models Dream-v0-7B-Base~\citep{ye2025dream} and LLaDA-8B-Base~\citep{nie2025llada}.\footnote{We do not compare with LLaDA-2.0~\citep{bie2025llada20scalingdiffusionlanguage} and SDAR-8B~\citep{cheng2025sdar} since their base models are not open-sourced.}

\paragraph{Reasoning capabilities.}
\ourmodel-8B-Base demonstrates performance competitive with autoregressive baselines, exhibiting particular strength in reasoning-intensive benchmarks for math, science, and code generation. 
Specifically, it outperforms Qwen3-8B-Base on advanced reasoning tasks such as MATH and GPQA, and achieves competitive performance on code generation. 
\ourmodel-8B-Base surpasses Dream-v0-7B-Base by a substantial margin, improving MATH performance from $39.6\%$ to $55.8\%$ and HumanEval from $57.9\%$ to $69.5\%$. 
As shown in Appendix~\ref{app:pretrain_ablation}, block diffusion consistently outperforms standard full-sequence diffusion models in reasoning performance when initialized from the same base model checkpoint.

\section{From Empirical Insights to \ourmodel-8B}
\label{sec:sft}
\subsection{Pilot Study: Block-Size Trade-offs}
\label{sec:pilot}
\begin{table*}[t]
\centering
\footnotesize
\setlength{\tabcolsep}{4pt}
\caption{Comparison of diffusion (LowConfidence) and autoregressive (AR) decoding strategies across varying training and inference block sizes on mathematical reasoning benchmarks.}
\label{tab:main}
\begin{tabular}{llcccccc}
\toprule
\multirow{2}{*}{\textbf{Training}} & \multirow{2}{*}{\textbf{Inference}} & \multicolumn{2}{c}{\textbf{AIME 2024}} & \multicolumn{2}{c}{\textbf{AIME 2025}} & \multicolumn{2}{c}{\textbf{MATH500}}  \\
\cmidrule(lr){3-4} \cmidrule(lr){5-6} \cmidrule(lr){7-8} 
& & \textbf{LowConfidence} & \textbf{AR} & \textbf{LowConfidence} & \textbf{AR} & \textbf{LowConfidence} & \textbf{AR}  \\
\midrule
\multirow{2}{*}{Block Size 4} 
 & Block Size 4 & 47.1 & 52.1 & 37.5 & 40.8 & 86.8 & 85.6 \\
 & Block Size 32 & 42.5 & 52.1 & 39.6 & 38.8 & 82.8 & 84.2 \\
\midrule
\multirow{2}{*}{Block Size 32} 
 & Block Size 4 & 20.0 & 29.2 & 24.2 & 22.5 & 78.8 & 78.2  \\
 & Block Size 32 & 42.5 & 48.3 & 21.3 & 33.3 & 80.2 & 83.4 \\
\midrule
\multirow{2}{*}{Curriculum 4$\rightarrow$32} 
 & Block Size 4 & 50.0 & 51.2 & 43.8 & 37.6 & 87.8 & 85.6  \\
 & Block Size 32 & 48.3 & 49.2 & 38.3 & 36.7 & 85.2 & 85.8   \\
\bottomrule
\end{tabular}
\end{table*}

To understand when block diffusion can preserve reasoning quality while increasing generation parallelism, we isolate the two endpoints of the block-size design space. Small blocks provide a stronger token-level causal scaffold but offer limited intra-block parallelism; Larger blocks expose more tokens to parallel denoising but place greater pressure on the model to maintain sequential fidelity within each block. We therefore conduct a pilot study on a filtered subset of PromptCoT~2.0~\citep{zhao2025promptcot}, retaining only mathematical examples with context lengths under 4096 tokens, and compare two fixed-granularity training strategies: Block Size~$4$ and Block Size~$32$.

\paragraph{Evaluation.} We assess these configurations under two decoding strategies. (1)~\textbf{LowConfidence}: following prior work~\citep{wu2025fastdllm,cheng2025sdar,bie2025llada20scalingdiffusionlanguage}, we unmask tokens whose confidence scores exceed a threshold of $\tau = 0.95$ at each denoising step. (2)~\textbf{Autoregressive}: As the canonical baseline, we adopt the standard left-to-right generation protocol, where at each step only the leftmost masked token is unmasked, strictly preserving the causal generation order. Evaluation is performed on AIME 2024, AIME 2025, and MATH500~\citep{hendrycksmath2021}.\looseness=-1

\paragraph{Granularity Alignment Under Diffusion Decoding.}
When evaluating under LowConfidence diffusion decoding, our results highlight a clear contrast between fine-grained and coarse-grained training. Direct training with large blocks (Block Size~32) suffers immense capability loss compared to its Block Size~4 counterpart, especially under fine-grained inference, where it reaches only $20.0\%$ on AIME 2024 and $24.2\%$ on AIME 2025 with Block Size~4 inference. This degradation demonstrates that coarse-block training forces the model to overfit to intra-block long-range dependencies, rendering it incapable of faithfully decomposing complex reasoning into fine-grained denoising steps. Conversely, Block Size~4 training exhibits remarkable robustness: it not only establishes strong performance in small-block settings but also maintains stable efficacy when evaluated with Block Size~32 inference suggesting that local structures learned from fine-grained block size training can organically support larger decoding blocks.\looseness=-1

\paragraph{Decoding Strategy: Autoregressive vs. LowConfidence.}
\citet{ni2026flexibilitytraparbitraryorder} reveal that arbitrary generation orders in full-sequence diffusion models often cause them to bypass high-uncertainty tokens, thereby collapsing the reasoning solution space, a phenomenon termed the \emph{flexibility trap}. To investigate whether block-wise diffusion encounters a similar trap, we compare decoding behaviors across granularities.\looseness=-1

Our evaluation reveals that training granularity strongly affects decoding resilience. With Block Size~4 training, the model achieves high absolute performance across both paradigms. While AR decoding provides a reliable performance ceiling (e.g., $52.1\%$ on AIME 2024), the model retains strong capabilities under diffusion decoding ($47.1\%$), implying it has successfully internalized robust local causal dependencies. In contrast, coarse-grained Block Size~32 training is substantially weaker under diffusion decoding on complex tasks (e.g., dropping to $21.3\%$ on AIME 2025), while strict AR decoding recovers part of the gap at $33.3\%$. This asymmetry indicates that direct large-block training fosters an over-reliance on parallel intra-block aggregation, compromising token-level sequential reasoning.

\subsection{Block-Size Curriculum Learning}
\label{sec:curriculum}

The pilot study exposes a granularity dilemma in block diffusion language models. To bridge this gap, we use a \textbf{block-size curriculum learning} framework for supervised fine-tuning our continually pretrained \ourmodel-8B-Base (\S\ref{sec:pretraining}).

Starting from this foundation, the curriculum first trains with Block Size~$4$ to acquire robust local causal dependencies and fine-grained reasoning patterns, and then exposes the model to larger blocks. In the pilot study, this second stage uses Block Size~$32$ to stress-test the block sizes. 

\begin{table*}[t]
\centering
\small
\caption{Performance comparison of open-source models on mathematics and code benchmarks.}
\label{tab:main_results}
\begin{tabular}{lcccc}
\toprule

\textbf{Model} & \textbf{Block Size}& \textbf{AIME 2024} & \textbf{AIME 2025} & \textbf{LiveCodeBench} \\
\midrule
\multicolumn{5}{l}{\textit{Autoregressive Baselines}} \\
DeepSeek-R1-Distill-Qwen-7B & - & 55.5 & 39.0 & 37.8 \\
MiMo-7B-RL & - & 68.2 & 55.4 & 50.7 \\
AceReason-Nemotron-1.1-7B & - & 72.6 & 64.8 & 52.1 \\
Qwen3-8B-Thinking & - & 76.0 & 67.3 & 52.8 \\
\midrule
\multicolumn{5}{l}{\textit{Diffusion Baselines}} \\
Dream-7B-Instruct & - & 0.0 & 3.3 & 9.7 \\
LLaDA-8B-Instruct & - &0.0 & 0.0 &  -\\
LLaDA-2.0-Flash (100B-A6B)& 32 & - & 60.0 & 41.0 \\
SDAR-8B-Chat & 4 & 13.8 & 10.4 & 16.4 \\
TraDo-8B-Instruct & 4 & 13.3 & 10.8 & 16.1 \\
TraDo-8B-Thinking & 8 & 35.5 & 17.5 & 14.2 \\
SDAR-30B-A3B-Sci & 4 & 73.4 & 59.2 & 29.0 \\
SDAR-30B-A3B-Sci & 8 &  52.5 &  40.4 & 6.4 \\
SDAR-30B-A3B-Sci & 16 & 44.2 & 39.6 & 3.2 \\
SDAR-30B-A3B-Sci & 32 & 47.1 & 23.3 &  2.6\\
\midrule
\multicolumn{5}{l}{\textit{Ours}} \\
\ourmodel-8B & 4 & 73.8 & 65.0 & 51.3 \\
\ourmodel-8B & 8 & 71.7 & 64.6 & 53.9 \\
\ourmodel-8B & 16& 71.7 & 62.9 & 53.6 \\
\ourmodel-8B & 32& 68.3 & 63.3 & 50.4 \\
\bottomrule
\end{tabular}
\end{table*}

\paragraph{Curriculum Learning Unifies Granularity Robustness.}
As shown by the curriculum rows in Table~\ref{tab:main}, the progressive curriculum (Block Size $4 \rightarrow 32$) emerges as the most robust diffusion-decoding performance across inference granularities. On AIME 2024, the curriculum model yields $50.0\%$ under Block Size~4 inference and $48.3\%$ under Block Size~32 inference, surpassing both fixed-block baselines. Similar trends are established on AIME 2025. Although the curriculum approach slightly trades off peak AR performance in some settings (e.g., $52.1\% \rightarrow 51.2\%$ on AIME 2024 compared to pure Block Size~4 training), it improves overall robustness. These findings suggest that gradual exposure to increasing block sizes preserves the fine-grained precision essential for small-block parallel denoising, while improving stability under large-block generation, reducing the need for maintaining multiple granularity-specific models.\looseness=-1

\subsection{\ourmodel-8B}
\label{sec:main_results}

Building upon the curriculum framework in Section~\ref{sec:curriculum}, we train our final model, \ourmodel-8B, on the complete PromptCoT~2.0 dataset~\citep{zhao2025promptcot}, encompassing approximately 4.8 million samples with context length 16384 tokens, spanning mathematical reasoning and code generation tasks. We initiate training with block size 4 and then transition to mixed-granularity training with block sizes randomly sampled from $\{4,8,16,32\}$. 

\paragraph{Baselines.} We compare against three categories of open-source models:
(1)~\textbf{Autoregressive baselines}: DeepSeek-R1-Distill-Qwen-7B~\citep{guo2025deepseekR1}, MiMo-7B-RL~\citep{xiaomi2025mimo}, AceReason-Nemotron-1.1-7B~\citep{chen2025acereason}, and Qwen3-8B-Thinking~\citep{yang2025qwen3};
(2)~\textbf{Full-attention diffusion baselines}: LLaDA-8B~\citep{nie2025llada} and Dream-7B~\citep{ye2025dream};
(3)~\textbf{Block diffusion baselines}: General-purpose instruct models, including LLaDA-2.0-Flash~\citep{bie2025llada20scalingdiffusionlanguage}, TraDo-8B-Instruct~\citep{wang2025revolutionizing}, and SDAR-8B-Chat~\citep{cheng2025sdar}, alongside reasoning-optimized variants with extended chain-of-thought capabilities, \mbox{TraDo-8B-Thinking}~\citep{wang2025revolutionizing} and SDAR-30B-A3B-Sci~\citep{cheng2025sdar}.\looseness=-1

\paragraph{Evaluation.} We assess mathematical reasoning through accuracy on AIME 2024 and AIME 2025, and measure code generation proficiency via pass@1 on LiveCodeBench (Oct 2024--May 2025)~\citep{jain2024livecodebench}. To assess architectural generalizability across varying block sizes, both SDAR-30B-A3B-Sci and our model are comprehensively evaluated at inference block sizes of 4, 8, 16, and 32. Remaining diffusion models are evaluated at their native training block sizes. All diffusion models are decoded using the LowConfidence strategy with threshold $\tau = 0.95$.

\paragraph{Comparison with Autoregressive Baselines.}

\ourmodel-8B achieves competitive performance against 8B-scale autoregressive reasoning models. Through continual pretraining and block-size curriculum learning, our model substantially narrows the gap with its autoregressive counterpart Qwen3-8B-Thinking. These results demonstrate that diffusion-based architectures, when properly trained, offer a viable alternative to autoregressive approaches for complex reasoning tasks.\looseness=-1

\begin{figure}[t]
    \centering
    \includegraphics[width=1.0\linewidth]{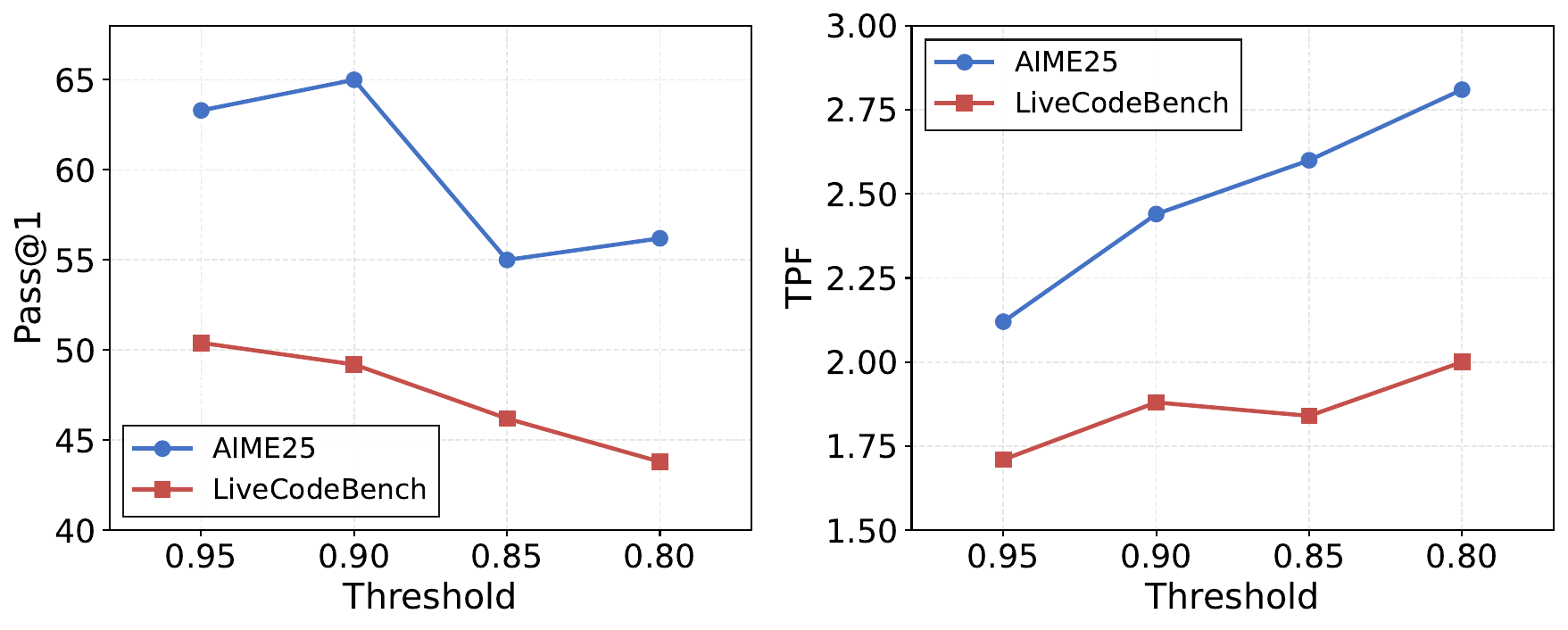}
    \caption{Effect of confidence threshold on reasoning accuracy and decoding efficiency (Block Size=32).}
    \label{fig:threshold}
\end{figure}

\paragraph{Comparison with Diffusion Baselines.}
Our model establishes a new performance frontier among open-source diffusion language models. Full-sequence diffusion language models such as LLaDA and Dream struggle to produce meaningful results on complex reasoning benchmarks. Among block diffusion approaches, \ourmodel-8B outperforms same-scale reasoning variants such as TraDo-8B-Thinking by a large margin, while surpassing substantially larger models including LLaDA-2.0-Flash (100B-A6B) and SDAR-30B-A3B-Sci across both mathematical and code benchmarks. The advantage is particularly pronounced on LiveCodeBench, where our model achieves a pass@1 score of 51.3\% compared with SDAR-30B-A3B-Sci's 29.0\%. 

\paragraph{Robustness to Inference Block Sizes.}
As shown in Table~\ref{tab:main_results}, \ourmodel-8B maintains stable performance across block sizes from 4 to 32, with only modest degradation on mathematical reasoning and consistently strong code generation. This resilience contrasts sharply with SDAR-30B-A3B-Sci, whose performance degrades substantially when inference block sizes exceed its training configuration. These results demonstrate that our curriculum learning framework decouples reasoning quality from inference block granularity, enabling greater generation parallelism without the severe accuracy penalties observed in prior block diffusion models.\looseness=-1

\begin{figure}[t]
    \centering
    \includegraphics[width=1.0\linewidth]{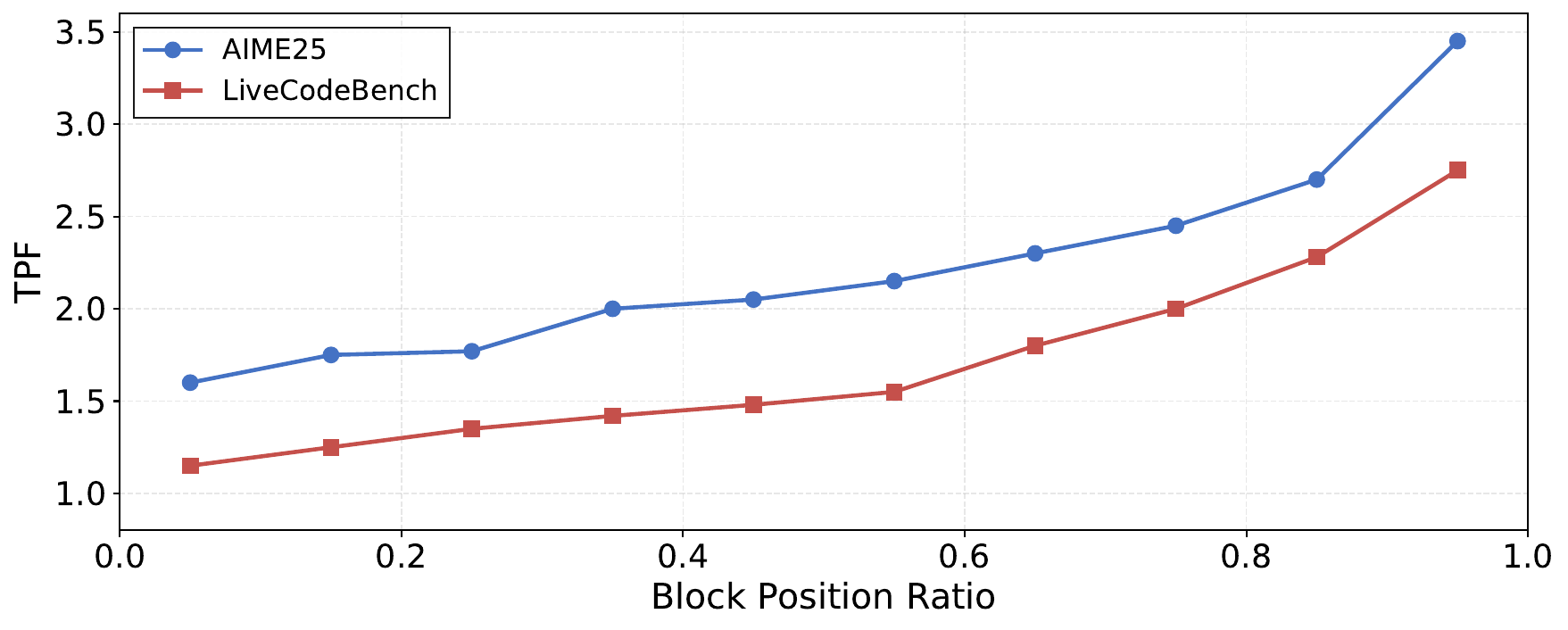}
    \caption{Tokens per forward pass across relative block positions under LowConfidence decoding ($\tau$ = 0.95).}
    \label{fig:tpf_block}
\end{figure}
\section{Analysis}
\subsection{Efficiency Analysis}

To better reflect the decoding efficiency, we measure tokens per forward pass (TPF) as the number of generated tokens divided by the total number of forward passes, counting both token-prediction passes and KV-cache recomputation steps.

\paragraph{Block Size.}
Figure~\ref{fig:overview} characterizes the accuracy--efficiency trade-off as the inference block size varies. TPF improves monotonically with larger block sizes across both benchmarks, indicating that coarser decoding configurations can effectively reduce the number of forward passes required for generation. At the same time, task accuracy remains relatively stable.

\paragraph{Threshold.}
Figure~\ref{fig:threshold} examines the effect of the confidence threshold on accuracy and TPF. Lowering the threshold permits more tokens to be decoded per forward pass, thereby improving TPF, yet risks accepting low-confidence predictions that compromise reasoning fidelity. Both benchmarks exhibit a general degradation trend as the threshold becomes more aggressive, with code generation showing particular sensitivity to token-level errors.

\paragraph{Thinking Process.}

We analyze decoding efficiency during the thinking and answering phases under LowConfidence decoding with a fixed confidence threshold of 0.95. Figure~\ref{fig:tpf_block} illustrates the average TPF within each block as a function of its relative position in the generated response.

TPF exhibits a monotonic upward trend on both AIME 2025 and LiveCodeBench benchmarks. In early-stage blocks (relative position $\approx 0.0$), the model commits only $\sim$1.2--1.6 tokens per forward pass, as the majority of tokens fall below the conservative threshold due to the scarcity of reasoning context. As generation progresses toward later stages (relative position $\approx 1.0$), TPF rises to $\sim$2.8--3.5, since the accumulated reasoning trace furnishes sufficient contextual information to support higher degrees of parallel token generation.

\begin{figure*}[t]
    \centering
    \includegraphics[width=0.85\linewidth]{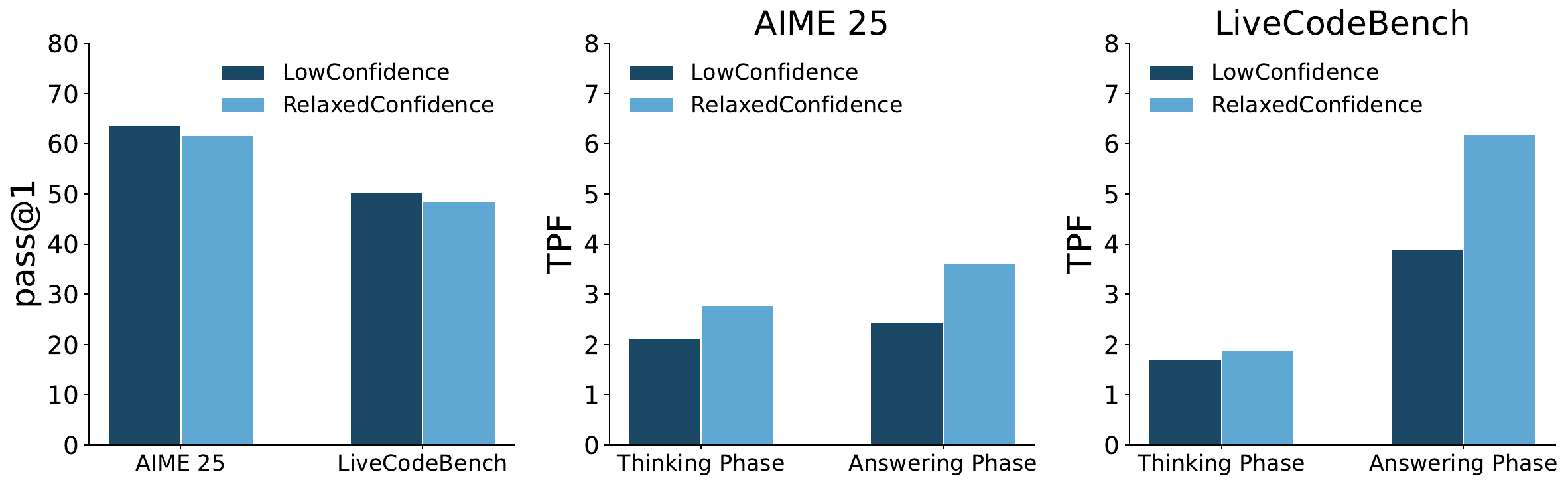}
    \caption{Pass@1 and throughput comparison between LowConfidence and RelaxedConfidence decoding.  We define the answering phase as the final solution-writing segment for AIME and the final code block for LiveCodeBench.}
    \label{fig:tpf_combined}
\end{figure*}

\subsection{Reasoning Acceleration Analysis}
Our step-by-step analysis reveals that standard confidence-threshold decoding is often overly conservative, as contiguous token spans typically stabilize before their individual confidence scores exceed the threshold~\citep{kong2025accelerating}. This raises the question of whether such implicitly stable tokens can be committed earlier without sacrificing accuracy. To investigate this, we introduce RelaxedConfidence as an analytical decoding strategy.

\paragraph{Formulation.}

For a masked token at position $i$, let $\mathcal{N}_r(i)=\{j:0<|i-j|\le r\}$ denote the valid positions within radius $r$, excluding $i$ itself. Let $\mathcal{D}$ be the set of already decoded positions, and define the reliable-position set $\mathcal{R}=\mathcal{D}\cup\{j:c_j\ge\kappa\}$, where $c_j$ is the model confidence for position $j$ and $\kappa$ is the neighbor-support threshold. With proximity weights $w_{i,j}=r-|i-j|+1$, we compute a spatial support score and position-specific commitment threshold as
\[
S_i =
\frac{
\sum_{j\in\mathcal{N}_r(i)\cap\mathcal{R}} w_{i,j}
}{
\sum_{j\in\mathcal{N}_r(i)} w_{i,j}
},
\quad
\tau_i = \tau - (\tau-\tau_{\min})S_i .
\]
Here $\tau$ is the standard LowConfidence threshold and $\tau_{\min}$ is the minimum threshold floor. A token at position $i$ is committed when $c_i\ge\tau_i$. Thus, isolated tokens with $S_i=0$ must satisfy the full threshold $\tau_i=\tau$, while well-supported tokens with $S_i=1$ can be committed at the relaxed threshold $\tau_i=\tau_{\min}$. This interpolation allows RelaxedConfidence to decode contiguous spans more aggressively when their local context is reliable, while preserving the full threshold for isolated positions.

\paragraph{Acceleration Without Degradation.}
Figure~\ref{fig:tpf_combined} summarizes the impact of RelaxedConfidence on reasoning quality and decoding throughput compared to the standard low-confidence re-masking baseline.
The leftmost panel reveals that RelaxedConfidence incurs only marginal degradation in accuracy on AIME25 and LiveCodeBench. This suggests that spatially informed threshold relaxation enables accelerated parallel decoding without compromising reasoning ability, likely because it primarily commits tokens that are well-supported by their local context.

\paragraph{Throughput Gains in Thinking and Answering.}
To localize where RelaxedConfidence accelerates decoding, we partition responses into a \textit{thinking} phase and an \textit{answering} phase.
During thinking, RelaxedConfidence raises TPF modestly---from $\sim$2.1 to $\sim$2.8 on AIME~25 ($+33\%$) and from $\sim$1.7 to $\sim$1.9 on LiveCodeBench ($+12\%$). Gains are substantially larger during answering: on AIME~25, TPF rises from $\sim$2.4 to $\sim$3.6 ($+50\%$); on LiveCodeBench, from $\sim$3.9 to $\sim$6.2 ($+59\%$). This reflects the more structured and lexically constrained nature of final responses, where contiguous token spans more readily reach implicit stability, allowing the relaxation criterion to commit longer coherent spans per pass.\looseness=-1

\section{Related Work}
\label{sec:related}

\subsection{Diffusion Language Models}
\label{sec:related-dlms}

Diffusion language models generate text through iterative denoising over discrete token vocabularies~\citep{austin2021structured,hoogeboom2021argmax,campbell2022continuous} or continuous representations~\citep{li2022diffusion,gong2022diffuseq,han-etal-2023-ssd,jo2025continuous}. Training objectives and parameterizations for discrete diffusion have been extensively refined~\citep{Zheng2023ARD,lou2023discrete,shi2025simplifiedgeneralizedmaskeddiffusion,zhao2024improving,sahoo2024simple}, with recent work scaling these models toward the modern LLM regime:  \citet{nie2025scalingmaskeddiffusionmodels} scaled to billion-parameter tasks; \citet{gong2025diffusionllm} adapted pretrained AR checkpoints to diffusion; \citet{nie2025llada} trained the 8B-parameter LLaDA from scratch; and \citet{ye2025dream} demonstrated competitive large-scale diffusion LMs initialized from AR checkpoints.

However, full-sequence diffusion struggles to maintain global coherence over long contexts, particularly for chain-of-thought reasoning~\citep{wang2025revolutionizing}. Earlier approaches \citep{han-etal-2023-ssd,wu2023ardiffusion,ye2024diffusion} introduced semi-autoregressive or position-dependent denoising mechanisms for text generation. Block diffusion~\citep{arriola2025blockdiffusion} formalizes this idea by partitioning a sequence into contiguous blocks with autoregressive factorization and denoising tokens within each block in parallel. This architecture interpolates between pure AR generation and full-sequence diffusion.
LLaDA2.0~\citep{bie2025llada20scalingdiffusionlanguage}, FastDLLM v2~\citep{wu2025fastdllmv2}, and EfficientDLM~\citep{fu2025efficientdlm} extended this paradigm to larger-scale settings but lack long chain-of-thought reasoning. T*~\citep{xia2026tstar} and TraDo~\citep{wang2025revolutionizing} studies progressive block scaling for masked diffusion models, increasing block size stage by stage to improve mathematical reasoning under larger blocks. Our work differs in treating block size as a deployment-time quality-efficiency variable and providing an in-depth analysis of the interplay between block size and decoding strategy.

\subsection{Reasoning with Language Models}
\label{sec:related-efficient-reasoning}

Reasoning in autoregressive language models has been substantially advanced by chain-of-thought techniques along with fine-tuning and reinforcement learning~\citep{jaech2024o1,guo2025deepseekR1,xiaomi2025mimo,yang2025qwen3}. 
Existing RL methods for diffusion language models spanning token-level likelihood surrogates~\citep{zhao2026d1,tang2025wd1,wang2025spg,gong2025diffucoder} and sequence-level objectives~\citep{zhu2025llada15, ou2025principled} have shown promise for improving reasoning, yet their evaluation remains confined to elementary math benchmarks such as GSM8K~\citep{cobbe2021training} and MATH~\citep{hendrycksmath2021}, leaving more complex real-world reasoning challenges such as AIME largely unexplored.
Previous block diffusion reasoners lag behind autoregressive counterparts~\citep{cheng2025sdar, wang2025revolutionizing} and face performance collapse when using large block sizes. Our proposed block-size curriculum mitigates the instability of direct large-block training, preserves the efficiency benefits of coarse-grained parallel decoding, and yields a strong open 8B block-diffusion reasoning model competitive with same-scale autoregressive reasoning models.

\section{Conclusion}

We present \ourmodel-8B, an open block-diffusion reasoning model that matches leading autoregressive baselines on math and code benchmarks. We show that fixed large-block training causes performance collapse while curriculum learning of block sizes stabilizes large-block reasoning. We further demonstrate that inference efficiency remains bottlenecked by conservative per-token commitment thresholds, and propose RelaxedConfidence, a context-aware decoding strategy that relaxes thresholds based on neighboring token confidence to accelerate parallel generation without compromising reasoning fidelity. We release our model and checkpoints to support future research on efficient diffusion-based reasoning.

\section*{Limitations}

While block-size curriculum learning effectively stabilizes large-block training and decouples reasoning quality from inference granularity, we do not explore dynamic or variable-length block sizes that respect semantic or syntactic boundaries, such as sub-problem delimiters, code blocks, or natural paragraph breaks. Allowing the model to adaptively choose block granularity rather than fixing it at predetermined widths could further push the Pareto frontier between decoding efficiency and modeling quality. We leave the investigation of boundary-aware, content-adaptive block partitioning to future work.

Due to computational constraints, our model is primarily limited to mathematical reasoning and code generation. We do not systematically assess the proposed curriculum on broader tasks such as tool use~\citep{yao2025tau} and coding agents~\citep{jimenez2024swe}. Generalizing block-size curriculum learning to these more diverse reasoning domains remains an important direction for future work.

%
\bibliography{custom}
\newpage
\appendix

\section{Implementation Details}
\label{app:impl_details}
\paragraph{Training.}
We extend Megatron-LM~\citep{shoeybi2019megatron} to support block diffusion training with a no-shift data loader~\citep{ni2025diffusion,fu2025efficientdlm} that uses input sequences directly as supervision targets. A noise injection module stochastically masks tokens while enforcing at least one masked token per block to prevent degenerate cases. The resulting interleaved sequences are processed via FlexAttention~\citep{dong2024flexattention}, which compiles the structured sparse attention pattern into optimized kernels, with training loss computed exclusively on masked positions.

In \S\ref{sec:pilot}, we train for 4 epochs in total. For the curriculum setting, the first 3 epochs use block size 4, followed by 1 epoch with block size 32.
In \S\ref{sec:main_results}, we first train with block size 4 for 3 epochs, then transition to mixed-granularity training with block sizes randomly sampled from $\{4, 8, 16, 32\}$. We intentionally preserve small block sizes in this final stage, as we observe that training solely with large blocks on longer thinking traces causes severe performance degradation during small-block inference.

\paragraph{Evaluation of Base Models.} 
We evaluate \ourmodel-8B-Base on a comprehensive suite of standard benchmarks spanning 
general knowledge, reasoning, mathematics, and code generation. 
\textit{General tasks} include MMLU~\citep{hendrycks2020measuring} for multi-subject understanding, 
ARC-Easy and ARC-Challenge~\citep{clark2018think} for commonsense reasoning, HellaSwag~\citep{zellers2019hellaswag} 
for sentence completion, PIQA~\citep{bisk2020piqa} for physical reasoning, WinoGrande~\citep{sakaguchi2021winogrande} 
for pronoun disambiguation, and RACE~\citep{lai2017race} for reading comprehension. 
\textit{Mathematical and scientific reasoning} is assessed through GSM8K~\citep{cobbe2021training} 
and MATH~\citep{hendrycks2020measuring} for mathematical problem-solving, and GPQA~\citep{rein2023gpqa} for 
graduate-level science questions. \textit{Code generation} capabilities are measured 
via HumanEval~\citep{chen2021evaluating} and MBPP~\citep{austin2021program}. We follow the evaluation framework in Dream-7B~\citep{ye2025dream} to evaluate our model. 

\paragraph{Evaluation of Reasoning Models.}
We adopt SGLang\footnote{\url{https://github.com/sgl-project/sglang}} as the inference engine. All baselines are reproduced in our codebase, except for Dream and LLaDA, whose results are taken from prior works~\citep{wang2025revolutionizing, xie2025dreamcoder}. The maximum generation length is set to 24K for both autoregressive and block diffusion baselines. In the pilot study (\S\ref{sec:pilot}), it is set as 8192 tokens. We set the temperature to 0.8 for autoregressive baselines and to 0 for diffusion baselines, as the latter employ threshold-based decoding. We run each baseline 8 times and report the average to reduce randomness.

\paragraph{RelaxedConfidence.} In RelaxedConfidence, we set the standard threshold to $\tau=0.95$, the neighborhood radius to $r=4$, and fix both the neighbor-support threshold $\kappa$ and the minimum relaxed threshold $\tau_{\min}$ at $0.7$.

\section{Continual Pretraining Ablation}
\label{app:pretrain_ablation}

We initialize both full-sequence and block diffusion from the same base model, Qwen3-8B~\citep{yang2025qwen3}, and continually pretrain each variant on 100B high-quality tokens drawn from the Olmo3 mid-training corpus. We evaluate both models on standard math and code reasoning benchmarks.

As shown in Table~\ref{tab:pretrain_ablation}, full-sequence diffusion lags behind block diffusion by a considerable margin on average across all evaluated tasks. These results demonstrate that block diffusion is substantially more effective for reasoning-intensive scenarios.

\begin{table*}[htbp]
\centering
\small
\caption{Comparison of full-sequence and block diffusion continually pretrained from Qwen3-8B on math and code reasoning benchmarks.}
\label{tab:pretrain_ablation}
\begin{tabular}{lcccc}
\toprule
Model & GSM8K & MATH & HumanEval & MBPP \\
\midrule
Qwen3-8B-Base & 86.5 & 52.7 & 65.8 & 68.8 \\
\midrule
\ourmodel-8B-Base & 83.5 & 51.2 & 62.8 & 60.4  \\
\quad w/o block diffusion  & 75.5 & 40.6 & 31.1 & 41.6 \\
\bottomrule
\end{tabular}
\end{table*}

\end{document}